\documentclass[conference]{IEEEtran}

\IEEEoverridecommandlockouts
\usepackage{cite}
\usepackage{amsmath,amssymb,amsfonts}
\usepackage{algorithmic}
\usepackage{graphicx}
\usepackage{pifont}
\usepackage{multirow}
\usepackage{booktabs}
\usepackage{textcomp}
\usepackage{xcolor}
\usepackage{makecell}
\usepackage{stfloats}
\def\BibTeX{{\rm B\kern-.05em{\sc i\kern-.025em b}\kern-.08em
    T\kern-.1667em\lower.7ex\hbox{E}\kern-.125emX}}
\begin{document}

\title{Transferring Vision-Language-Action Models to Industry Applications: Architectures, Performance, and Challenges\\

\thanks{This work has been submitted to the IEEE for possible publication. Copyright may be transferred without notice, after which this version may no longer be accessible. This work is supported by Liaoning Liaohe Lab with project number LLL24ZZ-02-01 and LLL24ZZ-02-02.}
}

\author{\IEEEauthorblockN{1\textsuperscript{st} Shuai Li}
\IEEEauthorblockA{\textit{Shenyang Institute of Automation} \\
\textit{Chinese Academy of Sciences}\\
Shenyang, China \\
lishuai1@sia.cn}
\and
\IEEEauthorblockN{2\textsuperscript{nd} Yizhe Chen}
\IEEEauthorblockA{\textit{Shandong Normal University} \\
Shandong, China \\
2023317064@stu.sdnu.edu.cn}
\and
\IEEEauthorblockN{3\textsuperscript{rd} Dong Li}
\IEEEauthorblockA{\textit{Shenyang Institute of Automation} \\
\textit{Chinese Academy of Sciences}\\
Shenyang, China \\
lidong@sia.cn}
\and
\IEEEauthorblockN{4\textsuperscript{th} Sichao Liu}
\IEEEauthorblockA{\textit{Department of Production Engineering} \\
\textit{Royal Institute of Technology (KTH)}\\
Stockholm, Sweden \\
sicliu@kth.se}
\and
\IEEEauthorblockN{5\textsuperscript{th} Dapeng Lan}
\IEEEauthorblockA{\textit{University of Chinese} \\
\textit{Academy of Sciences}\\
Beijing, China \\
dapengl@ieee.org}
\and
\IEEEauthorblockN{6\textsuperscript{th} Yu Liu*}
\IEEEauthorblockA{\textit{University of Chinese} \\
\textit{Academy of Sciences}\\
Beijing, China \\
liu.yu@ieee.org}
\and
\IEEEauthorblockN{7\textsuperscript{th} Zhibo Pang}
\IEEEauthorblockA{\textit{Department of Intelligent Systems} \\
\textit{Royal Institute of Technology (KTH)}\\
Stockholm, Sweden \\
zhibo@kth.se}
}

\maketitle

\begin{abstract}
The application of artificial intelligence (AI) in industry is accelerating the shift from traditional automation to intelligent systems with perception and cognition. Vision-language-action (VLA) models have been a key paradigm in AI to unify perception, reasoning, and control. Has the performance of the VLA models met the industrial requirements? In this paper, from the perspective of industrial deployment, we compare the performance of existing state-of-the-art VLA models in industrial scenarios and analyze the limitations of VLA models for real-world industrial deployment from the perspectives of data collection and model architecture. The results show that Pi0, after fine-tuning, achieves approximately 60\% success rate on simple grasping tasks, while the positional error in high-precision placing tasks reaches up to 2.2 cm and 12.4°. There is much room for performance improvement in complex industrial environments, diverse object categories, and high-precision placing tasks. Our findings provide practical insight into the adaptability of VLA models for industrial use and highlight the need for task-specific enhancements to improve their robustness, generalization, and precision. 
\end{abstract}


\section{Introduction}
Traditional control methods (e.g., PID control) demonstrate precise and reliable performance in the execution of a structured environment and repetitive tasks. Visual servoing can precisely align a screwdriver tip to M4-M8 screws, with average error of 0.8-1.3 mm~\cite{b1}. However, there are significant challenges when dealing with an unstructured environment characterized by complex, dynamic, and diverse tasks, such as inefficient human interactions, limited manufacturing tasks, and a lack of self-learning intelligence~\cite{b2}.

Vision-language-action (VLA) models are multimodal AI systems that integrate visual perception, language understanding, and action generation within a unified framework~\cite{b3, b4}. First, vision-language models (VLMs) pre-trained in Internet-scale data~\cite{b5, b6, b7} exhibit strong vision-language alignment capabilities, allowing open-vocabulary visual question answering in daily life scenarios~\cite{b8, b9, b10}. Building upon these pre-trained VLMs, VLA models can be further trained on large-scale robot data, equipping them with basic control capabilities. These models generate control commands conditioned on visual input and natural language instructions through autoregressive or diffusion-based policy heads~\cite{b11, b12, b13, b14, b15, b16, b17}. Besides, unlike daily life and laboratory scenarios, industrial environments are characterized by distinct technical indicators (e.g., illumination intensity $\geq 500\,\mathrm{lx}$, vibration amplitude $\pm 0.5\,\mathrm{mm}$). Fine-tuning with collected industrial task-specific data enables the transfer of pre-trained VLA models originally designed for daily-life scenarios into industrial scenarios. This adaptation allows VLA models to retain their inherent generalization capabilities, enabling them to handle tasks in partially unstructured environments to a certain extent. 

\begin{table}[h]
\caption{Comparison of review studies on VLA}
\centering
\small
\resizebox{0.95\linewidth}{!}{%
\begin{tabular}{lcccccc}
\toprule
\textbf{Survey} & \textbf{Year} & \textbf{General} & \textbf{Specific} & \textbf{VLA Coverage} & \textbf{Quantitative Evaluation} & \textbf{Applications} \\
\midrule
Ma et al. & 2024 & \ding{51} & \ding{55} &  & \ding{55} & \ding{55} \\
Sapkota et al. & 2025 & \ding{51} & \ding{55} & 80 & \ding{55} & \ding{55} \\
This Survey. & 2025 & \ding{51} & \ding{51} & 18 & \ding{51} & \ding{51} \\
\bottomrule
\end{tabular}
}
\label{tab:survey_comparison}
\end{table}

However, can the state-of-the-art VLA fulfill the industry requirement? To what extent must performance be improved to align with the anticipated objectives? These questions are still unclear. As shown in Table~\ref{tab:survey_comparison}, existing surveys focus on summarizing methods in general domains. VLA models are designed primarily for autonomy, intelligence, and adaptability. However, they still face limitations in terms of operational precision, real-time responsiveness, and system stability when deployed in industrial settings. These limitations can be attributed to three key factors:

\begin{itemize}
\item Lack of large-scale industrial data: Pre-trained VLMs fail to align domain-specific natural language instructions (e.g.,‘insert the shaft into the press-fit bearing') with corresponding visual inputs, limiting their out-of-distribution (OOD) generalization capabilities in industrial scenarios. 
\item High-precision requirements in an unstructured environment: Occlusions, diverse object categories, and complex spatial arrangements in an unstructured environment impose stringent demands on the control accuracy of VLA models. Although they can perform random picking within a single task, precision placing tasks remain a challenge.  
\item Computational resources and inference latency: The larger models typically require substantial memory and exhibit slow inference speeds. While action chunking has been introduced to improve inference efficiency, it inherently presents a trade-off: larger chunk sizes enhance speed but compromise fine-grained control accuracy.
\end{itemize}

In this study, we focus on evaluating the applicability and adaptability of VLA models in complex industrial environments. Specifically, we investigate the potential of VLA systems for use in real-world manufacturing settings, particularly under unstructured and dynamic conditions.

The main contributions of this study are as follows.
\begin{itemize}
\item We evaluated the performance of state-of-the-art VLA models in picking and placing tasks within industrial scenarios. 

\item We analyze the adaptability of VLA models to unstructured environments from two perspectives: dataset and model architecture. Furthermore, we also discuss potential directions to improve their robustness and task generalization.
\end{itemize}

\section{related work}

Unstructured environments characterized by complex, dynamic, and diverse tasks demand VLA models that exhibit generalization, control precision, and computational efficiency. Thus, we categorize prior works by their capabilities in high-precision requirements, real-time execution, and deployability.

\subsection{High-precision Requirements}
To meet high-precision requirements, VLA models require accurate imitation of demonstration data and fine-grained alignment between language and observations. 

\textbf{Visual perception:} Visual-language grounding and spatial understanding. Embodied Chain-of-Thought (ECoT) is a structured reasoning mechanism that trains VLA models to perform multiple steps of reasoning about plans, subtasks, motions, and visually grounded features such as object bounding boxes and end effector positions, before predicting robot action~\cite{b18}. ECoT improves the perceptual effectiveness of VLA models by explicitly guiding the alignment between subtasks and spatial features. EMMA-X~\cite{b19} is an embodied multimodal action model with a grounded chain of thought and spatial reasoning, allowing for finer task decomposition and significantly improving success rates over ECoT and OpenVLA. Both models transform language and observations into interpretable, structured, and controllable planning units. 3D-VLA~\cite{b20} unifies 3D perception, reasoning, and action with a generative world model. It enhances spatial understanding and localization in complex scenes by leveraging point cloud perception. Using natural language as a shared semantic representation, FuSe~\cite{b21} proposes a cross-modal finetuning approach to integrate heterogeneous sensory modalities, such as touch and audio, into pre-trained VLA models (e.g., Octo~\cite{b22} and PaliGemma~\cite{b9}), thus improving robustness in visually degraded conditions such as occlusion or low lighting.

\textbf{Language understanding:} Task decomposition and semantic reasoning. Through large-scale pretraining, VLM models acquire the ability to 1) align natural language instructions with visual context to enable semantic understanding of task goals; 2) identify and localize task-relevant entities (e.g., objects, spatial relations). RoboVLMs~\cite{b23} conducts over 600 systematic experiments across CALVIN, SimplerEnv, and real-world robot platforms such as Kinova Gen3, evaluating eight different VLM models. The results consistently show that more substantial vision-language alignment leads to better downstream control performance, with models such as KosMos~\cite{b24} and PaliGemma~\cite{b9} outperforming others. Beyond these, models such as Prismatic VLMs~\cite{b8}, Qwen2-VL~\cite{b25}, and Eagle-2~\cite{b10} are also commonly adopted as VLM backbones in recent VLA frameworks. Furthermore, the flexibility of VLM-based architectures supports modular VLA design, where VLMs can be paired with various policy heads, including autoregressive, diffusion, and hybrid, to predict action sequences. In summary, VLMs serve as the semantic backbone of VLA, mapping multimodal perception to middle-semantic and planning representation through their powerful alignment and generalization capabilities.

\textbf{Action generation:} Action head design and control accuracy. OpenVLA~\cite{b11} discretizes each action dimension into 256 uniformly spaced tokens based on quantile statistics and embeds these tokens directly into the LLaMA tokenizer. FAST~\cite{b12} improves on this by applying the Discrete Cosine Transform to compress action sequences, followed by Byte Pair Encoding to generate compact and information-dense tokens. Compared to OpenVLA, this representation enables faster training and higher precision in high-frequency tasks. Models such as Pi0~\cite{b14}, GR00T-N1~\cite{b16}, and GraspVLA~\cite{b17} use flow matching to align visual language conditions with continuous action trajectories, ensuring policy stability and precision. RDT-1B~\cite{b13} and CogACT~\cite{b15} employ probabilistic diffusion denoising models to model the distribution from visual language inputs to actions, enabling high-fidelity trajectory synthesis through multistep sampling. These continuous-space diffusion approaches generally outperform discrete tokenization methods in both representation quality and control accuracy. DiVLA~\cite{b26} adopts a 'reasoning-then-diffusion' framework, where an autoregressive module first performs task interpretation and subgoal decomposition, which is subsequently followed by a diffusion module responsible for fine-grained action generation. This design enhances both interpretability and execution accuracy by decoupling high-level semantic reasoning from low-level motion synthesis. HybridVLA~\cite{b27} advances this direction by integrating autoregressive and diffusion-based mechanisms within a unified vision language model. Through coordinated training and action ensemble strategies, it delegates diffusion modules to handle fine-grained motion control, while autoregressive modules are tasked with stage-level task planning. This complementary design enables each component to focus on its respective strength during execution.

\subsection{Model Efficiency and Real-Time Responsiveness}
Real-time responsiveness is essential for deploying VLA models in the real world, where latency directly impacts control stability. A series of recent works focus on model compression, lightweight architecture design, and adaptive inference strategies. SARA-RT~\cite{b28} introduces a robust self-adaptive attention mechanism that replaces traditional softmax attention with efficient linear attention through up-training, enabling large-scale VLA models such as RT-2 to achieve real-time inference without compromising performance. TinyVLA~\cite{b29} replaces the conventional autoregressive policy head with a diffusion-based policy head, enabling parallelizable and low-latency inference while maintaining high action precision. Evaluated for both simulated and real-world robotic tasks, TinyVLA demonstrates up to 20× lower latency and superior success rates compared to OpenVLA, all without relying on robot-specific large-scale pretraining. DeeR-VLA~\cite{b30} proposes a dynamic early-exit mechanism designed for multimodal LLMs. By dynamically distributing computation across time steps based on prediction confidence, it achieves 5–6× reductions in FLOPs and memory consumption with negligible performance degradation. In addition, diffusion-based models can also generate action sequences with large chunk sizes, balancing real-time and accuracy. 

\subsection{Deployability}
Deployability is a crucial consideration for the practical application of VLA models in industrial applications. We evaluated deployability from three dimensions: (1) open source model and code, (2) validation on real-world robot data, and (3) existence of real-world deployment demonstrations. Among existing works, the work stands out for its high level of openness and deployment on real robots, such as GR-1 and Mobile ALOHA.

\section{Architecture of VLA}
\begin{figure}[h]
\centering
\includegraphics[width=0.40\textwidth]{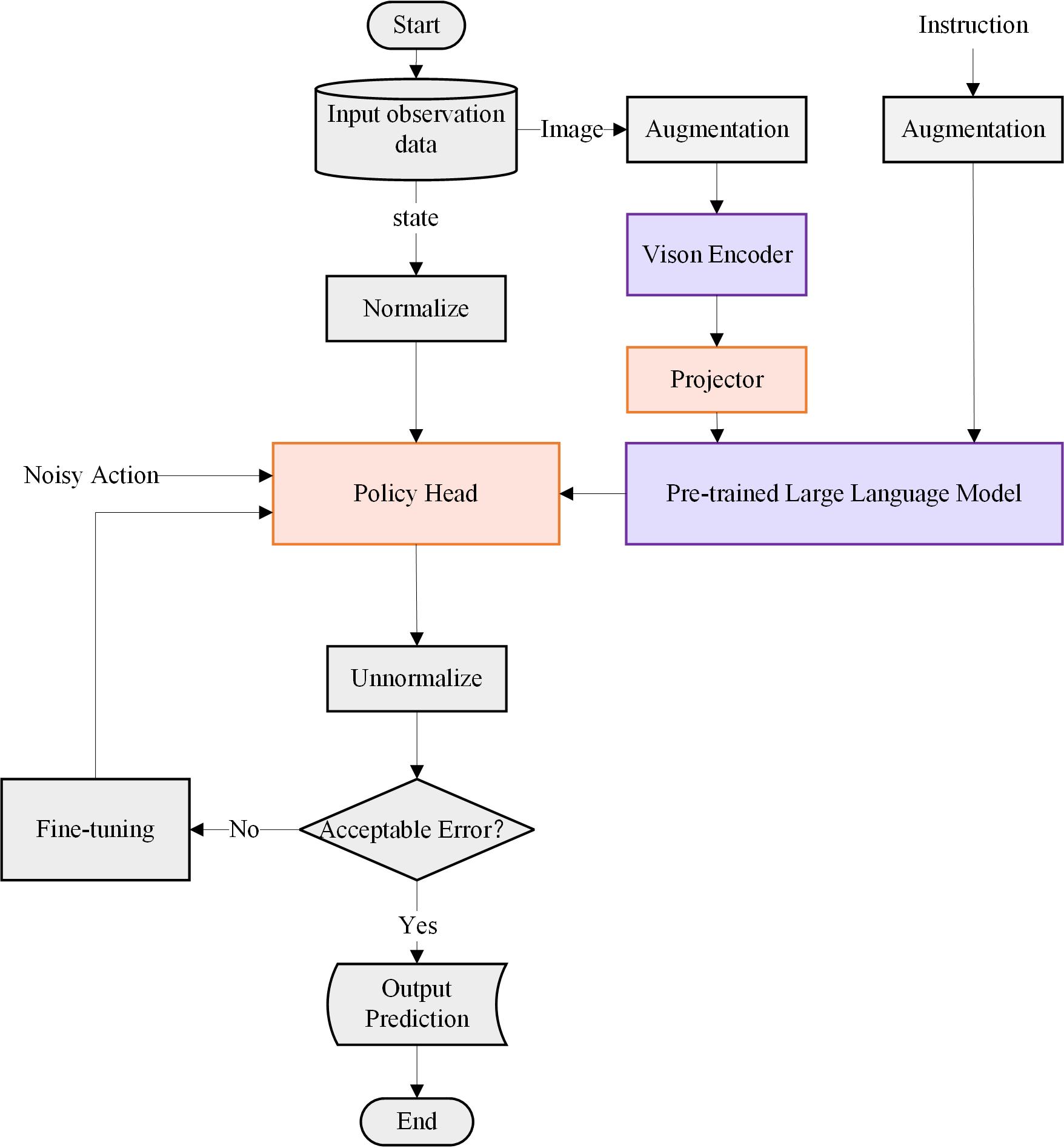}
\caption{Generalized technical framework for VLA.}
\label{fig}
\end{figure}

The existing reviews lack sufficiently detailed analyses of technical components, so we propose a generalized technical framework for VLA, as shown in Fig.~\ref{fig}. The framework can be interpreted as the disassembly and amalgamation of technical components within the prevalent methodological paradigms. Detailed insights into the technical components of various prevalent methods are delineated in Table~\ref{tab1}. Several primary technological modules encompass normalization and unnormalization, augmentation, a projector, pre-trained VLM, and policy heads. A comprehensive overview of each module is provided below.

\subsection{Normalization and unnormalization}

Normalization eliminates dimensional inconsistencies across different tasks or robotic platforms, thereby enabling the model to learn consistent distributions of actions and states. As illustrated in Fig.~\ref{fig}, normalization and unnormalization are applied to both actions and states, depending on the model design. Specifically, normalization maps raw physical values (e.g., joint positions, gripper poses) into standardized ranges, facilitating stable learning and generalization. Unnormalization reverses this transformation, converting model output back into task-specific or robot-specific physical units for execution. VLA models can be broadly categorized into two types based on whether proprioceptive state inputs are used: action-only normalization, used when the model does not explicitly take robot state as input; action-and-state normalization, used when both are fed into the policy network. Common normalization strategies include: Quantile normalization, Z-score normalization (standardization to zero mean and unit variance), and Min-Max normalization (scaling values to a fixed interval such as [–1, 1]).

\subsection{Augmentation}

Augmentation is applied to both visual observations and language instructions, enabling the VLM backbone to better handle perceptual variability and instruction diversity. Image augmentation includes standard techniques such as random cropping, color jittering, image corruption, and adding Gaussian noise. These techniques are critical to improving robustness against variations in camera pose, illumination, and background clutter, which frequently occur in real-world robotic deployments. Instruction enhancement typically involves prompt variation, paraphrasing, or template-based sampling, which helps the language model generalize across semantically equivalent but syntactically diverse instructions (e.g., `grasp the red block' vs. `pick up the red cube'). 

\subsection{Projector}

Visual observations are encoded by a vision encoder into a high-dimensional latent representation, while language instructions are embedded via pre-trained LLM tokenizers. However, these representations typically differ in both dimensionality and semantic space. The projector serves as a bridge between the vision encoder and the LLM and is designed to perform both dimension alignment and semantic mapping. A projector network, often implemented as a linear layer, MLP, or FusedMLP, allows visual tokens to be concatenated or interleaved with language tokens as input to the LLM or the policy heads. These networks are defined by the following equation, where image embeddings $\mathbf{e}_{\text{image}} \in \mathbb{R}^{L \times h_{\text{text}}}$ is output by a learned projector $\mathbf{F_{\psi}}$ 

\begin{equation}
\mathbf{e}_{\text{image}} = F_{\psi}(p_{\text{image}})
\end{equation}
where linear projector $\mathbf{F^{\text{linear}}_\psi}$ consists of a single linear layer; MLP projector $\mathbf{F^{\text{mlp}}_\psi}$ consists of two Linear layers and a GELU activation layer in between; FusedMLP projector $\mathbf{F^{\text{fused}}_\psi}$ comprises three linear layers interleaved with two GELU activation layers. In particular, RDT-1B uses linear projectors to align $\mathbf{e}_{\text{image}}$ and $\mathbf{e}_{\text{instruction}}$.

\subsection{Pre-trained VLM}
Formally, pre-trained VLM takes as input an image $\mathbf{x}_{\text{image}} \in \mathbb{R}^{H \times W}$ and language instruction $\mathbf{x}_{\text{instruction}}$. These inputs are then fed to the vision encoder and the pre-trained LLM.

\textbf{Vision Encoder:} Vision encoders are responsible for extracting rich spatial and semantic representations from raw visual observations. Common choices include ViT-based models such as DINOv2~\cite{b31} and SigLIP~\cite{b32}. Different vision encoders offer complementary strengths in visual representation. DINOv2, a self-supervised ViT-based model, excels at capturing fine-grained spatial details and low-level geometric features, making it suitable for tasks requiring precise spatial reasoning and object localization. In contrast, SigLIP, a contrastive vision-language pre-training model, emphasizes semantic alignment and extracts high-level conceptual features that are strongly correlated with language instructions. Vision encoders are defined by the following equation.

\begin{equation}
p_{\text{image}} = V_{\omega}(x_{\text{image}})
\end{equation}
where $V_{\omega}$ represents the vision encoders, $p_{\text{image}} \in \mathbb{R}^{L \times h_{\text{vision}}}$ could be the patch features output by DINOV2 and SigLIP.  

\textbf{Language Encoder/Pre-trained LLM:} LLMs serve as the core of reasoning and decision making in VLA architectures. They interpret natural language instructions, align them with visual representations, and generate either structured task representations or direct control actions. Common choices include LLaMA, Gemma, Qwen2-VL, and SmolLM2, which are typically used in frozen or lightly fine-tuned configurations. The LLM takes as input the tokenized language instruction embedding $\mathbf{e}_{\text{instruction}}$ along with the visual embeddings $\mathbf{e}_{\text{image}}$. This process is defined by the following equation.

\begin{equation}
f_{\text{gen}} = \text{LM}_\theta\left([\mathbf{e}_{\text{image}}; \mathbf{e}_{\text{instruction}}]\right)
\end{equation}
where $\text{LM}_\theta$ represents the LLM, $\mathbf{e}_{\text{image}}$ and $\mathbf{e}_{\text{instruction}}$ are concatenated. 

\subsection{Policy heads}
The policy head serves as the final component in VLA, mapping the fused language vision representations $f_{\text{gen}}$ to robot actions. Depending on the type of action space and the task requirements, different architectural variants and training objectives have been proposed. There are three main architectural types: autoregressive, diffusion, and hybrid models. 

\textbf{Autoregressive:} Autoregressive-based policy heads typically adopt a De-tokenizer structure, treating action prediction as a next-token prediction task over discretized tokens. These heads are optimized with cross-entropy (CE) loss. 

\textbf{Diffusion:} Diffusion-based policy heads operate in continuous action space and predict actions by denoising a noisy trajectory input. It includes different architectures, such as the DiT architecture, the Gemma architecture, and the scaleDP. These heads are optimized with Mean Squared Error (MSE) loss or flow-matching loss.

\textbf{Hybrid:} Depending on the architecture, LLM-dominant designs integrate diffusion modeling into the autoregressive prediction of the next token within a single LLM. In contrast, diffusion-dominant designs use a diffusion model as the main trajectory generator, with the LLM providing semantic conditioning or latent planning inputs. These heads are typically optimized with a combination of MSE and CE loss, aligning continuous action prediction with a discrete planning structure.

\begin{table*}[ht]
\caption{Constituents of diverse prevalent methods}
\begin{center}
\small
\resizebox{\linewidth}{!}{%
\begin{tabular}{lcccccccccccc}
\toprule
\multirow{2}{*}{\textbf{Method}} & \multirow{2}{*}{\textbf{Normalize}} & \multicolumn{2}{c}{\textbf{Augmentation}} & \multicolumn{2}{c}{\textbf{Projector}} & \multicolumn{2}{c}{\textbf{Pre-trained VLM}} & \multirow{2}{*}{\textbf{Policy Head}} & \multirow{2}{*}{\textbf{Unnormalize}} & \multirow{2}{*}{\textbf{Loss}} & \multirow{2}{*}{\textbf{Chunk size}} & \multirow{2}{*}{\textbf{Frequency (hz)}}\\
\cmidrule(lr){3-4}
\cmidrule(lr){5-6}
\cmidrule(lr){7-8}
 & & \textbf{Image Augmentation} & \textbf{Instruction Augmentation} & \textbf{Image} & \textbf{Text} & \textbf{Vision Encoder} & \textbf{Language Encoder/Pre-trained LLM} & & & & \\
\midrule

OpenVLA & Quantile &  &  & Linear, MLP, FusedMLP &  & DinoV2, SigLIP & Llama2-7B & De-Tokenizer & Quantile & CE &  & 5\\
RDT-1B &  & \makecell[l]{color jittering, image corruption,\\ and Gaussian noise} & GPT-4-Turbo & Linear & Linear & SigLIP & T5-XXL & DiT &  & MSE & 64 & 381\\
Pi0 & Z-score &  &  & Linear &  & SigLIP &  Gemma (2.6B) & Gemma (300M) & Z-score & Conditional Flow Matching  & 50 & 50\\
Pi0-Fast & Z-score &  &  & Linear &  & SigLIP &  Gemma(2.6B) & De-Tokenizer & Z-score & CE  &  & 50\\
CogACT& Quantile & \makecell[l]{random crop, random brightness,\\ random contrast, random saturation,\\ and random hue} &  & Linear, MLP, FusedMLP &  & DinoV2, SigLIP &  Llama-2-7B & DiT & Quantile & MSE  & 16 & 80\\
GR00T-N1& Min-max &  &  & MLP &  & SigLIP-2 &  SmolLM2 & DiT & Min-max & Flow Matching & 16 & 120\\
DiVLA& Z-score &  &  GPT-4o  & MLP &  & SigLIP & Qwen2-VL & ScaleDP & Z-score & MSE+CE  & 16 & 82\\
HybridVLA& Quantile &  &  & Linear, MLP, FusedMLP &  & DinoV2, SigLIP &  Llama-2-7B & De-Tokenizer, MLP & Quantile & MSE+CE  &  & 9.4\\
\bottomrule
\end{tabular}
}
\label{tab1}
\end{center}
\end{table*}

\subsection{Action space}
The design of the action space directly impacts the trade-off between control precision and learning efficiency. Joint space and task space control are the two dominant action modes for controlling robot arms within the robot learning literature~\cite{b33}. Furthermore, the action space can be further divided into absolute and relative control modes. Both are defined by the following equation.

\begin{equation}
relative \mathcal {A}_e = \left[ \Delta x, \Delta y, \Delta z, \Delta \phi, \Delta \theta, \Delta \psi, g \right]
\end{equation}

\begin{equation}
absolute \mathcal {A}_J = \left[ J_1, ... ,  J_n, g \right]
\end{equation}
where $\Delta x, \Delta y, \Delta z$ are the relative translation offsets of the end effector, $\Delta \phi, \Delta \theta, \Delta \psi$ denote the rotation changes. $\mathcal{A}_J \in \mathbb{R}^n$, where $n$ is the number of joints in the robot arm. $g$ indicates the gripper’s open/close state. 

In unstructured environments, absolute joint angle control ensures high precision and repeatability by avoiding inverse kinematics (IK) and its associated issues. However, it has lower training efficiency and limited generalization. In contrast, relative end-effector control offers better sample efficiency and generalization but depends on IK and lacks the precision and safety required in industrial settings.

\section{Evaluation of VLA}

We aim to evaluate the robustness and generalizability of VLA models in real-world industrial environments. To do this, we design a structured benchmark comprising three industrial scenarios, focusing on the models’ ability to perform random object picking and precision placing. 

\subsection{Experiment setups}
\begin{figure}[h]
\centering
\includegraphics[width=0.4\textwidth]{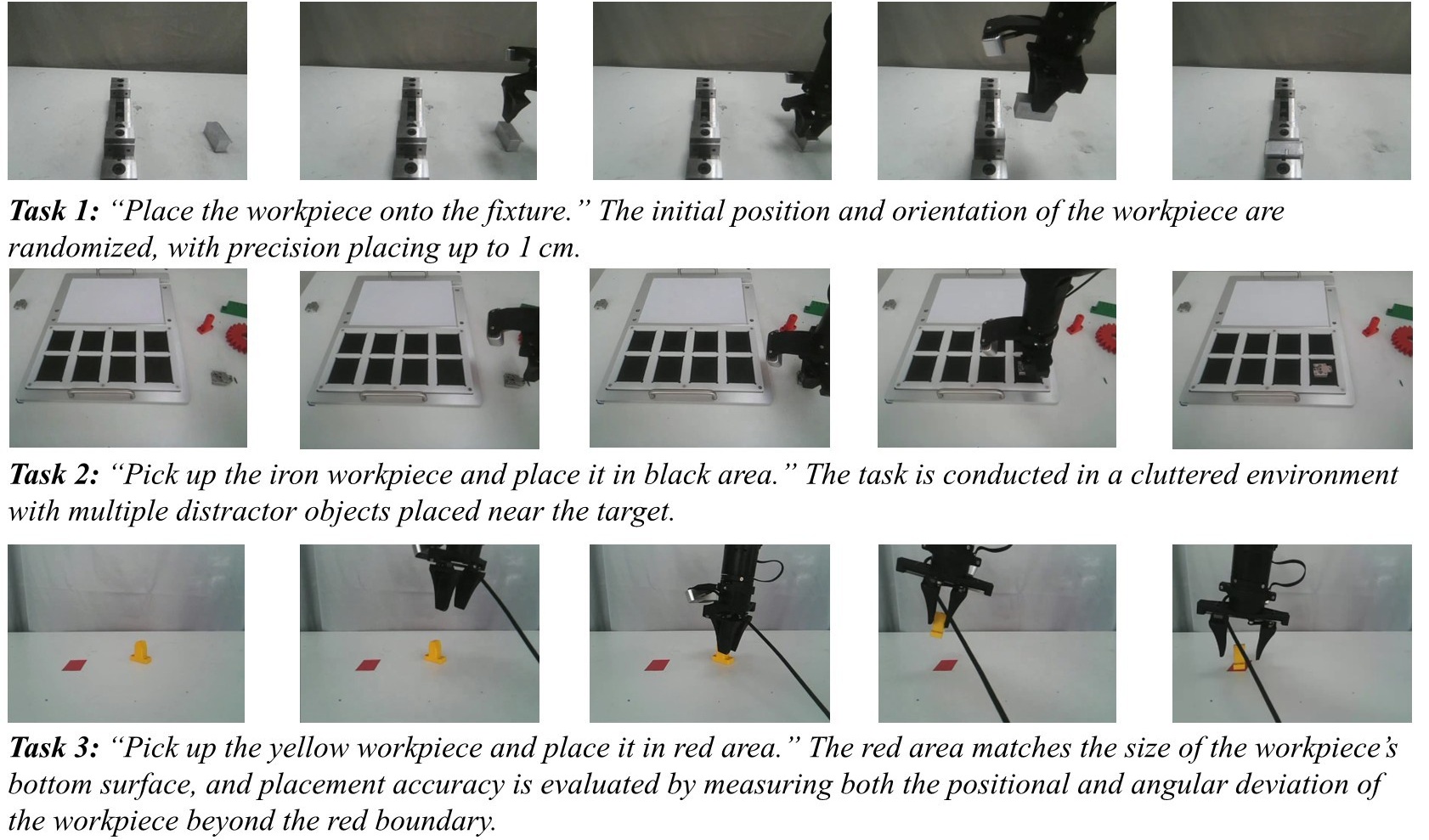}
\caption{Task definition and visualization.}
\label{fig2}
\end{figure}

\textbf{Task:} We consider four types of perturbation commonly encountered in industrial deployments: visual occlusion, camera jitter, object pose randomisation, and object diversity. An illustration of each task, detailed definitions, and visualizations are provided in Fig.~\ref{fig2}.

\textbf{Data:} We fine-tune the VLA model using episodes collected from real-world industrial scenarios, and 100 episodes are used for each task individually. 

\textbf{Method:} To evaluate VLA, we select Pi0 in robotic foundation models from deployability and real-time. As OpenVLA and CogACT rely solely on third-person demonstrations, their performance under occlusion conditions is limited due to the lack of egocentric visual input.

\textbf{Metric \& Hardware:} We employ the success rate (\%) as our main metric, which is calculated by dividing the successful trials by the total number of trials. The model is fine-tuned on an NVIDIA H20 96GB GPU for 10 hours using Ubuntu 20.04 and ROS 1 noetic. All tests are performed on the Mobile ALOHA dual-arm robot. Every trial incorporates object pose randomization and diverse object categories, 10 trials in total.

\subsection{Results analysis}
In the zero-shot setting, none of the evaluated VLA models successfully completed the grasping task. Table~\ref {tab:quantitative-results} provides a quantitative assessment of the effects of visual occlusion and camera jitter on task performance. All results were obtained after fine-tuning the models in real-world demonstrations. H-O and W-O denote the occlusion of head and wrist cameras, respectively, where 30\% of the pixels in the entire image are randomly masked. H-J refers to the head camera jitter. Performance degrades significantly under head and wrist camera occlusion (H-O, W-O), especially in Task 2, suggesting that visual occlusion has a more pronounced impact under object diversity. The relatively stable results under H-J across tasks imply the model’s robustness to moderate head camera jitter. However, the average error (2.2 cm, 12.4°) highlights a substantial potential for improvement when deploying Pi0 in unstructured environments.

\begin{table}[h]
\caption{Quantitative results}
\centering
\small
\resizebox{0.95\linewidth}{!}{%
\begin{tabular}{l cccc|cccc|cc}
\toprule
\multirow{2}{*}\textbf{Method} 
& \multicolumn{4}{c|}{\textbf{Random Pick Up}} 
& \multicolumn{4}{c|}{\textbf{Precision Place}} 
& \multicolumn{2}{c}{\textbf{Error}} \\
\midrule
\cmidrule(lr){2-5}
\cmidrule(lr){6-9}
\cmidrule(lr){10-11}
& \textbf{} & \textbf{H-O} & \textbf{W-O} & \textbf{H-J}
& \textbf{} & \textbf{H-O} & \textbf{W-O} & \textbf{H-J}
& \textbf{Distance (cm)} & \textbf{Orientation (°)} \\
Pi0-task1 & 60 & 50 & 60 & 60 & 40 & 20 & 20 & 20 &  &  \\
Pi0-task2 & 40 & 30 & 30 & 40 & 40 & 30 & 20 & 30 &  &  \\
Pi0-task3 & 50 & 40 & 30 & 40 & 50 & 20 & 20 & 30 & 2.2 & 12.4 \\
\bottomrule
\end{tabular}
}
\label{tab:quantitative-results}
\end{table}

\section{Discussions}
The deployment of VLA models in industrial settings introduces unique challenges due to the unstructured and dynamic nature of these environments. Based on the empirical results in Section IV, we revisit two key questions that directly reflect the gap between current VLA capabilities and industrial requirements.

Q1: Why do VLA models struggle to generalize to unstructured industrial environments? Most existing VLA models are designed for general-purpose or household manipulation tasks, focusing on broad generalization over precision. Industrial environments exhibit unique challenges such as visual occlusions, camera motion jitter, and diverse object arrangements, which are rarely addressed in existing architectures. For example, they lack explicit mechanisms to handle partial observability or visual disturbances and do not incorporate feedback loops to correct actions based on execution outcomes. Moreover, the limited availability of demonstration data and the scarcity of vision-language aligned datasets result in poor zero-shot transferability of VLA models to new robot arms and task layouts. Our experiments indicate that successful adaptation typically requires hundreds of domain-specific episodes for fine-tuning. 

Q2: Why do VLA models not yet satisfy industrial control requirements? In terms of action space design, the use of absolute joint angle control is more suitable for high-precision tasks compared to relative end-effector control, which relies on inverse kinematics and introduces instability in constrained workspaces. However, the inference latency of large VLA models (often below 10 Hz) poses challenges for real-time industrial deployment. These delays lead to unstable execution, including motion jitter and lag, which severely affect task reliability in industrial deployments. Moreover, current VLAs are highly sensitive to the distribution of training data. Differences in robot arm, joint configuration, or frequency between platforms require extensive data collection and retraining, which undermines general-purpose deploymentability. 

\section{Conclusions}

In this study, we evaluated state-of-the-art VLA models in real-world industrial scenarios to assess their applicability. While current models show the ability to perform randomized object picking under occlusion and jitter, their placing accuracy remains below industrial standards, with errors typically at the centimeter level. Our analysis attributes the limitations of VLA models to meet industrial requirements (high accuracy, real-time responsiveness, deployability) to limited high-quality industrial data, high computational demands, and insufficient attention to deployment constraints in model design. Despite some task adaptability, these models lack generalization across embodiments. Future work should focus on improving perceptual robustness, enabling real-time feedback, and developing lightweight architectures to bridge the gap toward industrial deployment.

\vspace{12pt}
\end{document}